# Deep Transfer Learning for Identifications of Slope Surface Cracks

Yuting Yang, Gang Mei*
School of Engineering and Technology, China University of Geosciences (Beijing), China
Email: yuting.yang@cugb.edu.cn; gang.mei@cugb.edu.cn

**Abstract**

Geohazards such as landslides have caused great losses to the safety of people's lives and property, which is often accompanied with surface cracks. If such surface cracks could be identified in time, it is of great significance for the monitoring and early warning of geohazards. Currently, the most common method for crack identification is manual detection, which is with low efficiency and accuracy. In this paper, a deep transfer learning framework is proposed to effectively and efficiently identify slope surface cracks for the sake of fast monitoring and early warning of geohazards such as landslides. The essential idea is to employ transfer learning by training (a) the large sample dataset of concrete cracks and (b) the small sample dataset of soil and rock masses cracks. In the proposed framework, (1) pre-trained cracks identification models are constructed based on the large sample dataset of concrete cracks; (2) refined cracks identification models are further constructed based on the small sample dataset of soil and rock masses cracks. The proposed framework could be applied to conduct UAV surveys on high-steep slopes to realize the monitoring and early warning of landslides to ensure the safety of people's lives and property.

# List of Abbreviations

| | |
|---|---|
| Fig | Figure |
| VGG | Visual Geometry Group Network |
| ResNet | Residual Network |
| ViT | Vision Transformer |
| DenseNet | Densely Connected Convolutional Networks |
| CNN | Convolutional Neural Networks |
| DCNN | Deep Convolution Neural Networks |
| GPU | Graphics Processing Unit |
| OS | Operating System |
| CPU | Central Processing Unit |
| CPU RAM | Central Processing Unit Random Access Memory |
| CUDA | Compute Unified Device Architecture |
| ACC | Accuracy |
| UAV | Unmanned Aerial Vehicle |

# 1 Introduction

Geological disasters (Geohazards) frequently occur in the world, and various geological disasters have caused great losses to the safety of people's lives and properties (Fan et al. 2019; Ma et al. 2020; Mei et al. 2020). The occurrence of geological disasters such as landslides, debris flow, rock collapse, surface subsidence, surface collapse, surface cracks, and earthquakes, etc., is often accompanied by the occurrence of surface cracks of soil and rock masses.

Especially for high-steep slopes that are more prone to geological disasters such as rock collapse and landslides (as illustrated in Fig. 1), a series of cracks will be formed in the early stage of the occurrence of geological disasters. The cracks caused by the collapse are illustrated in Fig. 2. (source: [www.news.sohu.com](www.news.sohu.com) and [www.heraldnet.com](www.heraldnet.com)). Before the rock collapse occurs, there are cracks on the top of the slope, and the cracks at the rock collapse area will gradually expand, and new cracks will appear at the foot of the slope over time (Ma et al. 2020; Mei et al. 2020).

Therefore, monitoring and identifying cracks on the top of a rock collapse slope can effectively provide an early warning of collapse and reduce the loss of human life and property caused by collapse [2-3]. Landslide cracks are illustrated in Fig. 2. Landslide cracks are an important sign accompanying landslides, which can be divided into tensile cracks, shear cracks, bulging cracks, and fan-shaped cracks (Ma et al. 2020; Mei et al. 2020). In the early stage of the formation of the landslide, intermittent cracks will appear on the rear edge of the landslide (Wang et al. 2020), and the length of the cracks tends to be constant. With continuous cracks appearing on the trailing edge and the length of the cracks showing an expanding trend, it indicates that the landslide is slowly intensifying (Du et al. 2021; Lian et al. 2020).

As the landslide slides down, a shear zone will be formed between the landslide and the parent body, and shear cracks will appear. The soil and rock masses distributed in the front of the landslide body will also form open cracks due to the uplift of the soil and rock masses hindered by the slide of the landslide body. At the same time, the spread of the landslide tongue to both sides will also form fan-shaped cracks(Du et al. 2016). The change of landslide cracks reflects the development process and extent of the landslide. Therefore, the identification of cracks at different positions of the landslide is of great significance for the early monitoring and warning of the landslide (Cheng et al. 2021).

Currently, the more commonly used crack detection methods are still mainly manual detection. However, there are several problems in manual detection, including detection blind spots in the detection process, high fieldwork intensity, low work efficiency, the subjective impact of detection results, and low accuracy(Cao et al. 2020; Yang et al. 2018).

To address the problems arising in manual detection, various methods based on image identification have been developed(Cao et al. 2020; Huang and Zhang 2012). However, those methods based on image identification are easily affected by environmental factors such as light, shadow, background, etc., and the computing of the later image-based identification methods is often very complicated, and the computational efficiency is also low. The reliability of the detection result of the crack picture with complex background is unsatisfactory(Cha et al. 2017a; Liu et al. 2019).

In recent years, deep learning has achieved good performance in object identification tasks and has the advantages of high parallelism, good robustness, and strong generalization ability. Deep learning does not need to design artificial feature extraction, especially in image classification; the accuracy of the trained deep learning network models is significantly higher than that of traditional machine learning approaches(Chadaram and Yadav 2020; Liu and Meng 2005; Zhang et al. 2020).

At present, many deep learning models with higher accuracy, generalization ability, and robustness are produced. Applying them to crack detection can effectively improve the efficiency of crack identification and make crack detection more efficient (Jang et al. 2019; Jiewen et al. 2020). Commonly used identification models include LeNet5 (LeCun et al. 1998), AlexNet (Krizhevsky et al. 2012), GoogLeNet (Szegedy et al. 2015; Szegedy et al. 2016), VGG (Simonyan and Zisserman 2015), ResNet (Deep Residual Network) (He et al. 2016), DenseNet (Dense Connected Convolutional Network) (Huang et al. 2017), MobileNets (Andrew G. 2017), etc. Recently, the Vision Transformer (ViT) (Alexey Dosovitskiy 2020) employed the Transformer model to image identification is proposed and widely used.

There is much research work that has been conducted to identify cracks using deep neural networks. For example, Zhang et al. (Zhang et al. 2016) employed the deep learning method for road crack detection. The deep convolutional neural network was used to perform crack identification on 500 3264×2448 images of strong noise environments (vehicles, tree shadows, etc.) collected by smartphones, which broadens the application scope of deep learning crack identification. Zhang et al. (Zhang et al. 2019) proposed an automated road crack detection method for three-dimensional asphalt pavement, which is an efficient detection framework based on a convolutional neural network (CrackNet).

Moreover, Cha et al. (Cha et al. 2017b) proposed a damage assessment method based on deep learning. This method is more robust and accurate in the detection of concrete cracks than traditional image detection methods. Chen et al. (Chen and Jahanshahi 2018) proposed the NB-CNN network framework, which fused convolutional neural network (CNN) and Naïve Bayes data. The framework aggregated each frame of information in the video for data fusion to further perform crack detection. The overall performance and robustness of the detection system have been improved.

In addition, Dorafshan (Dorafshan et al. 2018) et al. compared six traditional edge detection methods with Deep Convolution Neural Networks in terms of crack detection effects. DCNN shows greater advantages in terms of calculation time and accuracy, indicating the superiority of DCNN in crack detection. Maeda et al. (Maeda et al. 2018) used the smartphone in the car to obtain pictures to establish a dataset and successfully detected 8 road damages by training the convolutional neural network damage detection model. And compared the accuracy and running speed of GPU server detection and smartphone detection, and developed a smartphone application that has been put into use.

However, currently, there is little research work focusing on the use of deep learning methods for the identification of slope surface cracks, especially for the surface cracks of high and steep slopes. This is because the deep learning methods have a high dependence on the amount of data, and a large amount of training data is required for deep learning model training. However, it is difficult to obtain crack images of high and steep slopes, and the data sample size is small. In this case, there are difficulties in the construction of deep learning models.

To address the above problems, in this paper, a deep transfer learning framework is proposed to effectively and efficiently identify slope surface cracks for the sake of early monitoring and warning of geohazards such as landslides.

First, seven deep neural network models are used to identify the concrete cracks that are easier to obtain in the dataset, and the pre-trained crack identification models are constructed. Then, the transfer learning strategy is exploited by combining the small sample soil and rock masses crack dataset made by ourselves with the large sample concrete crack dataset. The pre-trained deep neural network models are used to identify soil and rock masses cracks. Finally, the deep learning models for the identification of slope cracks are obtained, which are expected to be used in slope monitoring and early warning of geological disasters to ensure the safety of people's lives and properties.

The rest of this paper is organized as follows. Section 2 describes the proposed deep learning framework in detail. Section 3 analyzes the results obtained by this framework. Section 4 discusses the advantages, applicability, and disadvantages of the proposed framework, as well as possible future work. Section 5 concludes the paper.

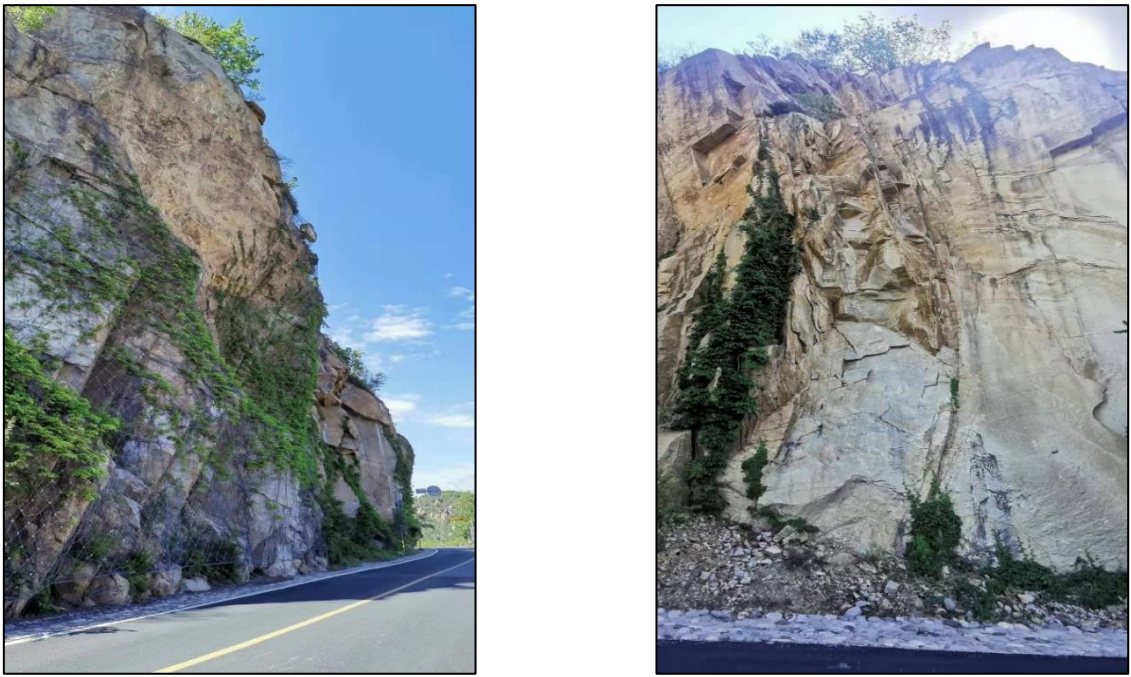

**Fig. 1** High and steep slopes

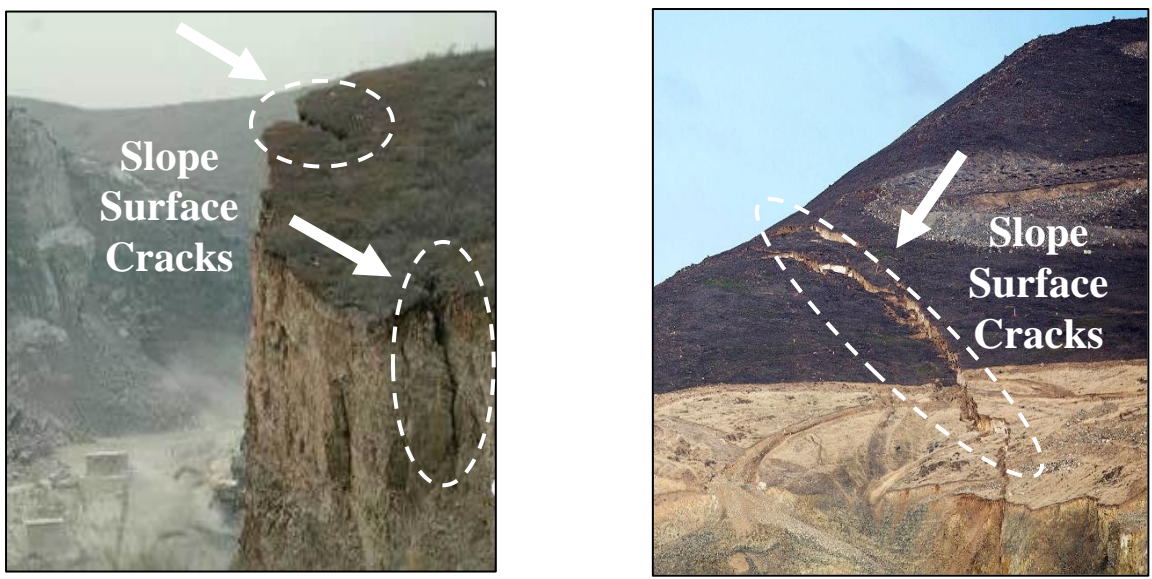

**Fig. 2** Real-world examples of slope surface cracks

# 2 Methods

## 2.1 Overview

In this paper, we propose a deep transfer learning framework to effectively and efficiently identify slope surface cracks for the sake of early monitoring and warning of geohazards such as landslides.

First, the deep learning models are pre-trained for the identification of slope surface cracks, and the models are compared and analyzed to screen out the optimized model. Then, the strategy of transfer learning is utilized. A small sample dataset of surface cracks of soil and rock masses is established and is combined with a large sample dataset of concrete cracks. The pre-trained models are applied to the identification of soil and rock masses slope surface cracks, thereby constructing the deep learning models for the identification of slope surface cracks.

The workflow of the proposed deep transfer learning framework is illustrated in Fig. 3.

**Step 1**: Collecting and processing the concrete crack dataset, and carrying out the pre-training of the deep learning models for the identification of slope surface cracks.

(1) Collecting and making a large sample dataset of concrete cracks.

(2) Using convolutional neural network models (e.g., LeNet5, AlexNet, LeNet5 model using InceptionA and InceptionE modules, ResNet18, MobileNet) and Vision Transformer to identify concrete cracks. Establishing a series of pre-trained deep learning models for identifying cracks in soil and rock masses slopes.

(3) Enhancing the data of each model. Comparing and analyzing the accuracy and calculation efficiency of each model. Analyzing the application range of each model, and screening out the optimized model to prepare for the identification of soil and rock masses slope surface cracks.

**Step 2**: Sorting out and labeling the dataset of surface cracks of soil and rock masses. Using the transfer learning strategy to obtain refined deep learning models for the identification of slope surface cracks.

(1) For the difficult-to-obtain surface cracks of soil and rock masses datasets, collecting, sorting, and labeling relevant data by ourselves. Designing and producing corresponding surface cracks of soil and rock masses small sample datasets and applying them to the identification of soil and rock masses slope surface cracks.

(2) Employing the strategy of transfer learning, a large sample dataset of concrete cracks and a series of pre-training models established are applied to the identification of soil and rock masses slope surface cracks. Establishing the deep learning models for the identification of surface cracks in the soil and rock masses slopes, which can be applied to the identification of surface cracks in the slope to serve the fast monitoring and early warning of geological disasters.

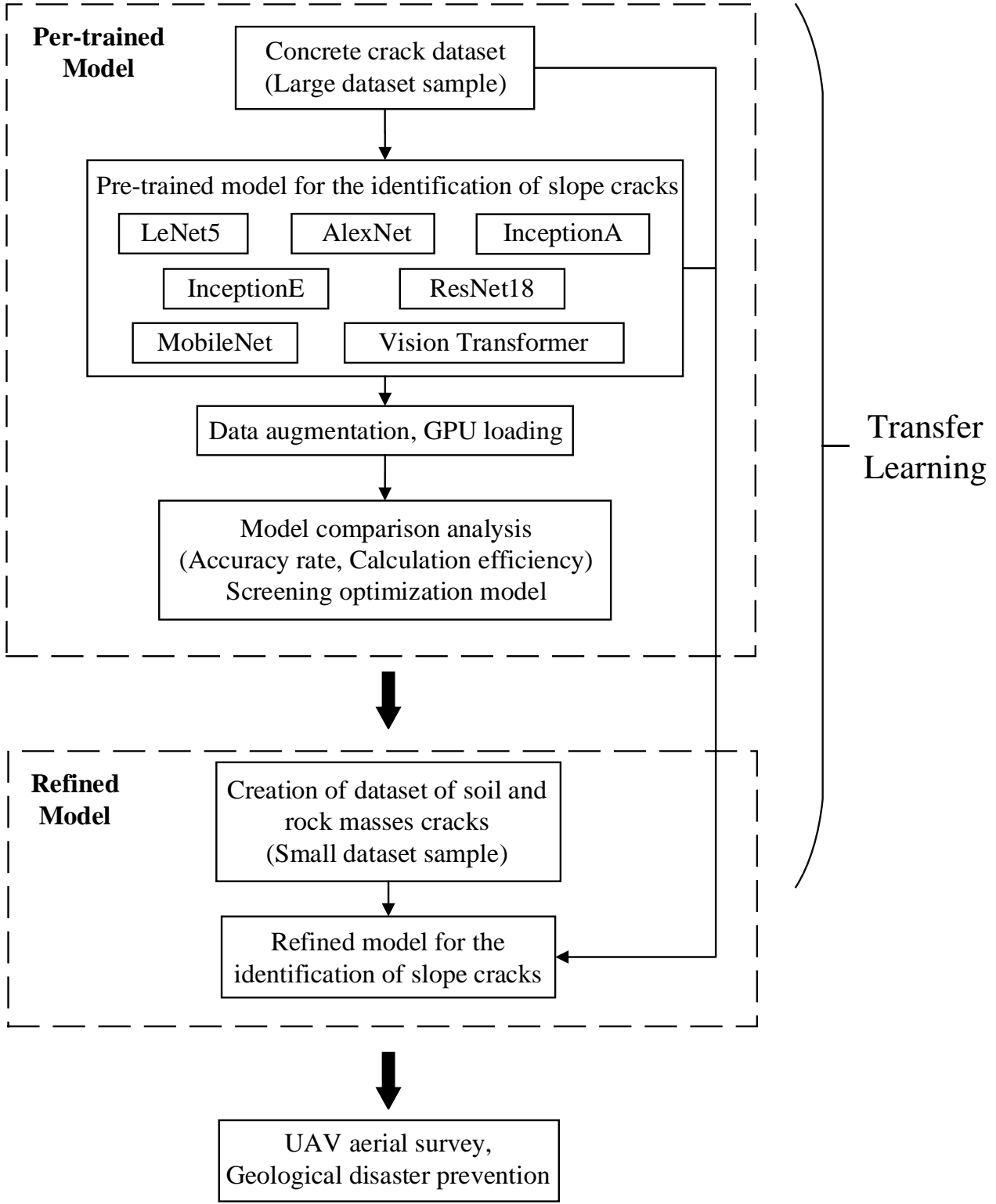

**Fig. 3** Workflow of the proposed deep learning framework for the identification of slope surface cracks

## 2.2 Step 1: Construction of Pre-trained Deep Learning Models for the Identification of Slope Surface Cracks

### 2.2.1 Data Collection and Cleaning

In this paper, we collect concrete crack images through publicly available datasets. The concrete images with cracks and the concrete images without cracks are collected separately. We crop the images to a uniform size, such as 227×227 size, remove the images with excessive interference, and finally save the same number of concrete images with and without cracks. 80% of them are randomly selected as the training set and 20% as the test set. Finally, the required concrete crack dataset is obtained.

### 2.2.2 Data Augmentation

Data augmentation refers to the process of performing a series of transformation operations on a limited dataset to generate new data, reduces network overfitting, and improves the generalization ability of the training models. In this paper, we employ data augmentation to address the problem of insufficient data in the dataset and expand the amount of data.

Data augmentation includes random flip transformation, random rotation transformation, scale transformation, random cropping, and center cropping transformation, as well as brightness, contrast, and hue transformation, as illustrated in Fig. 4. In this paper, the above-mentioned data augmentation methods are used in combination, through the `torchvision.transforms.Compose()` function, a data augmentation operation is performed before each data acquisition, which greatly expands the content of the dataset.

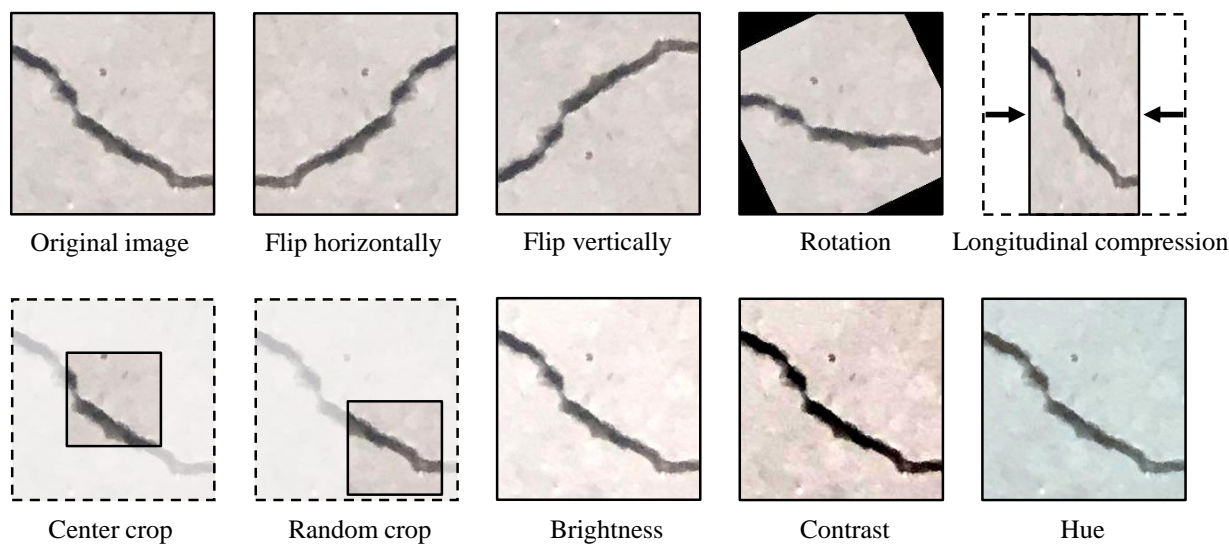

**Fig. 4** Comparative illustrations before and after data enhancement

## 2.2.3 Pre-trained Model Construction

The pre-trained models are constructed applying the following six deep neural network models. Finally, a total of seven deep neural network models are built, including LeNet5, AlexNet, InceptionA, InceptionE, ResNet, MobileNet, Vision Transformer. Each of the models is briefly described as follows.

(1) LeNet5: The most primitive image classification model LeNet5 is used for crack image identification. The LeNet5 model has a simple structure and a small number of network layers, but the image needs to be scaled to 32×32 first, and then trained on the LeNet5 model.

(2) AlexNet: It is the first convolutional neural network model that has attracted everyone's attention. Compared with LeNet, AlexNet has a deeper network structure and uses ReLU instead of Sigmoid as the activation function. A series of optimization measures have been taken to achieve higher accuracy.

(3) Inception module: GoogLeNet first employed the Inception module, enhanced the function of the convolution module, and further deepened the network depth. Later, there are InceptionV2, InceptionV3, and InceptionV4 network structures. Because the GoogLeNet network is too large, only the InceptionA and InceptionE modules in InceptionV3 are applied to LeNet respectively to further deepen the depth of the LeNet network and improve the accuracy.

(4) ResNet: The residual module Residual Block is introduced to alleviate the problem of gradient vanishing caused by the network being too deep. Using the ResNet18 network structure, compared to the network structure using the Inception module, the calculation speed is significantly improved.

(5) MobileNet: It is a lightweight deep neural network model, suitable for mobile terminals, embedded devices, and other devices that have the low computing power and require speed and real-time performance. The MobileNetV1 network structure is employed, and the deep separable convolution module is used to reduce the calculation parameters and the amount of calculation.

(6) Vision Transformer (ViT): The Transformer network model is now widely used in the fields of NLP and CV. It could achieve good results and high accuracy. In this paper, the Vison Transformer model is employed to crack identification, and the effect of crack image identification is further tested.

## 2.3 Construction of Refined Deep Learning Models for the Identification of Slope Surface Cracks

### 2.3.1 Dataset Construction

As the data of surface cracks on high and steep slopes are difficult to obtain, and the soil and rock masses cracks of the slope are also surface cracks of soil and rock masses. The surface cracks of soil and rock masses data are collected by ourselves and used as the identification of soil and rock masses slope surface cracks dataset. The obtained crack images of the soil and rock masses are sorted, cropped, and identified by ourselves, and the images of the soil and rock masses with surface cracks and the image of the soil and rock masses without surface cracks of the same size are respectively obtained. The number of images with cracks and images without cracks are equivalent. 80% of them are randomly selected as the training set and 20% as the test set to construct an image dataset of cracks in the slope to identify the cracks in the soil and rock masses.

## 2.3.2 Refined Model Construction

Deep learning has been widely used because it can learn advanced features of data. However, deep learning requires sufficient data to complete training and has a strong dependence on the amount of data. For many tasks, in reality, it is impossible to obtain a large number of trainable datasets, so transfer learning methods can be employed.

Transfer learning solves the problem of insufficient high-quality training data in the target field by transferring knowledge from the existing source field to the target field (Shao et al. 2015; Zhuang et al. 2021). In this way, the corresponding problems can be solved even in the case of a lack of data, which reduces the data dependence of deep learning. The basic principles of transfer learning are illustrated in Fig. 5.

When the amount of data is small, it is difficult to accurately classify the data. After employing similar problems for auxiliary training, although there will still be a certain deviation in the classification of the data, it can complete the classification task well. For example, for the classification of five-pointed stars and circles. After transfer learning, the task of classifying six-pointed stars and circles can be realized.

In this paper, we employ the idea of transfer learning, use the concrete cracks dataset, which is similar to the surface cracks of soil and rock masses to carry out the deep learning model pre-training. Constructing the pre-trained deep learning models. Then combining the small sample surface cracks of soil and rock masses dataset made by ourselves and the large sample concrete crack dataset to obtain an optimized dataset. The deep learning model is trained again, and finally, the refined deep learning models employed to the identification of soil and rock masses slope surface cracks are obtained.

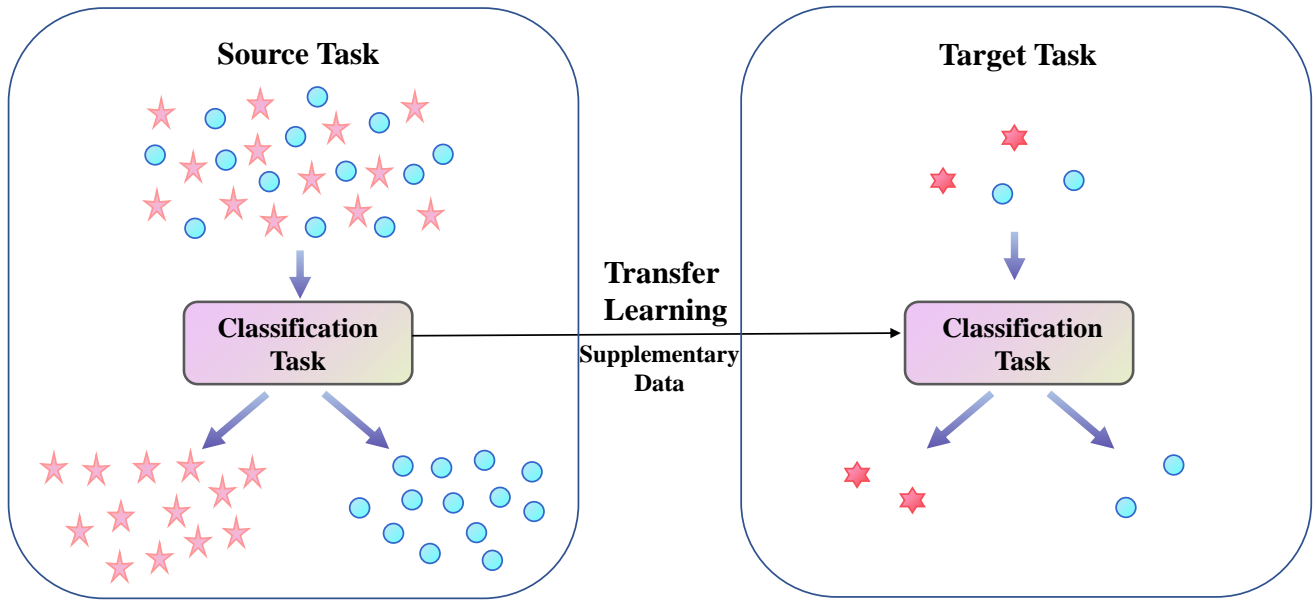

**Fig. 5** A simple illustration of transfer learning

# 3 Results

## 3.1 Experimental Environment

The software and hardware environment configurations used in this paper are listed in Table 1 and Table 2, respectively.

Table 1 Software environment configurations used in this paper

| Software | Details |
| --- | --- |
| OS | Windows 10 Professional |
| Programming language | Python |
| Deep learning framework | PyTorch |
| Dependent library | Torch、Torchvision、CUDA、PIL etc. |

Table 2 Hardware environment configurations used in this paper

| Hardware | Details |
| --- | --- |
| CPU | Intel Xeon Gold 5118 CPU |
| CPU Frequency (GHz) | 2.30 |
| CPU core | 48 |
| CPU RAM (GB) | 128 |
| GPU | Quadro P6000 |
| GPU memory (GB) | 24 |
| CUDA cores | 3840 |
| CUDA version | v9.0 |

## 3.2 Experimental Data

### 3.2.1 Pre-trained Model Dataset

In this paper, we first obtained the public concrete crack image dataset (source: www.kaggle.com) for deep learning model pre-training. There are a total of 40,000 images in the dataset, the image size is 227×227, and there are 20,000 images with and without cracks, respectively. 80% of images with cracks and images without cracks are randomly selected as the training set, and the remaining 20% are used as the test set. Therefore, the training set has a total of 32,000 images, of which there are 16,000 images with cracks and 16,000 images without cracks. The test set has a total of 8,000 images, of which there are 4000 images with cracks and 4000 images without cracks. The background images of some concrete with cracks and concrete without cracks in the dataset are illustrated in Fig. 7 and Fig. 8.

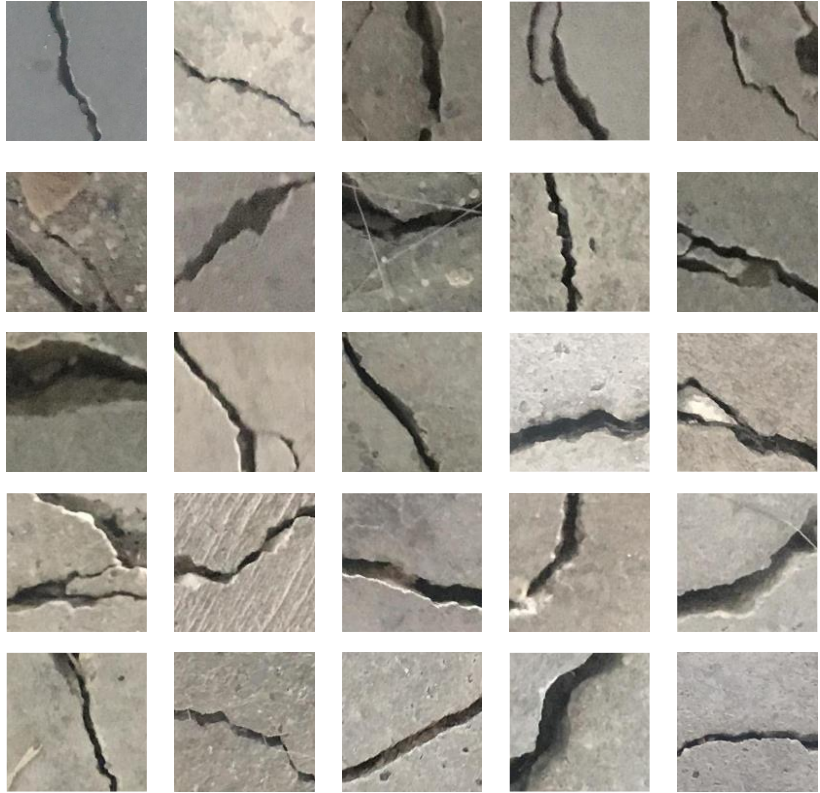

**Fig. 6** The concrete crack dataset that contain the images of the cracks

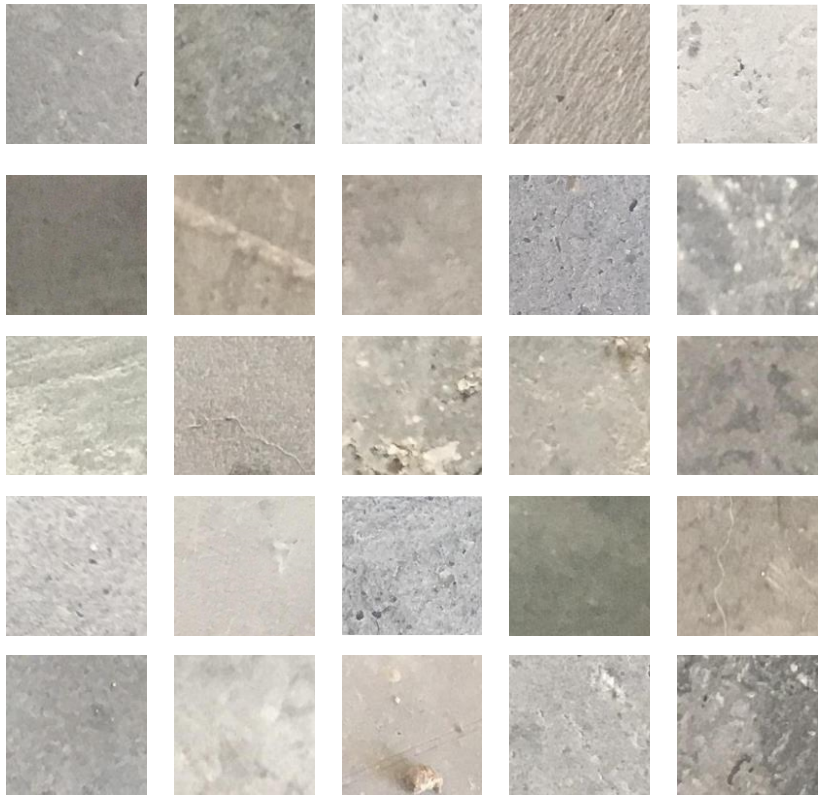

**Fig. 7** The concrete crack dataset that does not contain the images of the cracks

## 3.2.2 Refined Model Dataset

Since there is little related research work on the identification of surface cracks of soil and rock masses, the surface cracks of soil and rock masses dataset required for the construction of the refined model cannot be directly obtained. Thus, the surface cracks of soil and rock masses dataset need to be produced by ourselves. In this paper, we collect, sort out and crop the crack images of soil and rock masses, produce a smaller dataset for the identification of cracks in the soil and rock masses. The dataset has a total of 400 images, and the size of each image is adjusted to the same size as the original dataset: 227×227. Among them, 200 are images of soil and rock masses with surface cracks, and 200 are images of soil and rock masses without surface cracks. 80% of them are randomly selected as the training set of the rock and soil crack recognition dataset, and 20% as the test set. Therefore, there are a total of 320 images in the training set, among which there are 160 images with cracks and 40 images without cracks. There are a total of 80 images in the test set, of which there are 40 images with cracks and 40 images without cracks. Some images of the dataset are illustrated in Fig. 9 and Fig. 10.

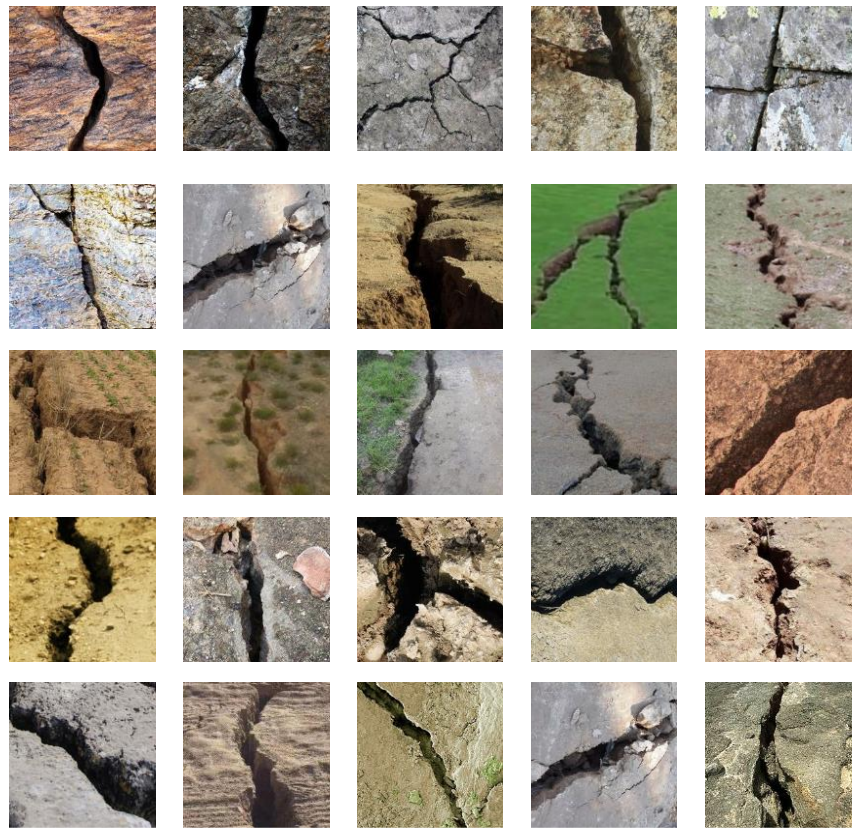

**Fig. 8** The dataset of surface cracks of soil and rock masses that contains images of cracks

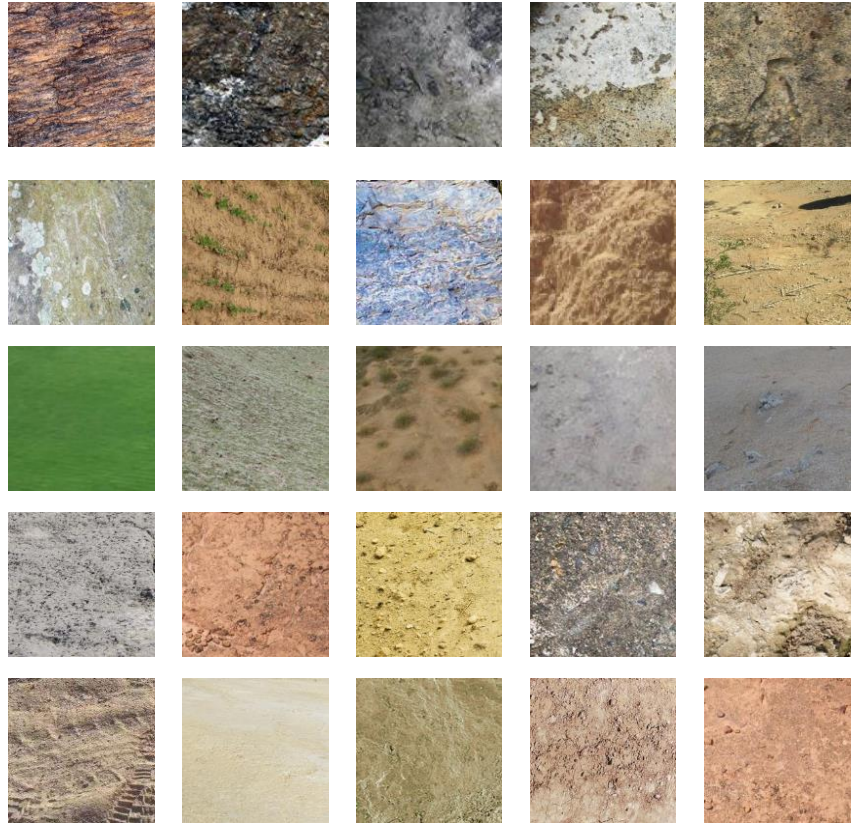

**Fig. 9** The dataset of surface cracks of soil and rock masses that does not contain images of cracks

## 3.3 Test Results and Analysis of the Pre-trained Deep Learning Models for the Identification of Slope Cracks

The deep neural networks used in this paper include LeNet5, AlexNet, the LeNet5 network model improved by Inception V3's Inception A and Inception E modules (referred to as InceptionA and InceptionE in the following for easy identification), ResNet18, MobileNet, and Vision Transformer. The loss function used in the model is `torch.nn.CrossEntropyLoss()`, and the optimizer is `torch.optim.SGD()`. The comparison and analysis of each model in terms of accuracy and efficiency are as follows.

3.3.1 Accuracy

When the data augmentation is not performed, the accuracy comparison of each model under 10 epochs is illustrated in Fig. 11, and the results are listed in Table 3. From Fig. 11 and Table 3, it can be seen that before data augmentation, the accuracy of each model is higher. Except for the slightly lower accuracy of Vision Transformer, the accuracy of other models can reach more than 99%. Moreover, the accuracy of the model does not fluctuate much in ten epochs, and the model is relatively stable. Vision Transformer has the lowest accuracy rate among the seven models, but it can also reach more than 94%, indicating that Vision Transformer has a broad space for future applications in the field of computer vision. Take the maximum accuracy of each model in 10 epochs for comparison, and the columnar analysis chart is illustrated in Fig. 12.

From the comparative analysis in Fig. 12, it can be seen that, before the data augmentation, AlexNet has the highest accuracy, reaching 99.800%. Except for Vision Transformer, LeNet5 has the lowest accuracy. This is due to the relatively simple structure of LeNet and the relatively deep model network. The accuracy of InceptionA and InceptionE improve based on LeNet5 is higher than that of LeNet5. This is due to the use of the Inception module to further deepen the network model. The accuracy of the ResNet18 network model can also reach 99.750%. Because its depth is the highest among the seven network models, its accuracy is also higher. However, due to the small dataset and fewer training images, ResNet18 failed to give full play to its Function, the accuracy rate is slightly lower than that of AlexNet.

Table 3 Accuracy of crack identification before data augmentation

| epoch | LeNet5 | AlexNet | InceptionA | InceptionE | ResNet18 | MobileNet | VisionTransformer |
|---|---|---|---|---|---|---|---|
| 1 | 98.600 | 99.312 | 98.513 | 97.588 | 95.875 | 98.013 | 94.263 |
| 2 | 98.812 | 99.375 | 98.875 | 99.000 | 99.562 | 99.025 | 94.588 |
| 3 | 98.912 | 99.575 | 99.237 | 99.263 | 99.638 | 99.400 | 94.388 |
| 4 | 99.000 | 99.688 | 99.263 | **99.275** | 99.200 | **99.638** | 94.513 |
| 5 | 98.800 | 99.125 | 99.312 | 98.862 | 99.688 | 98.950 | **94.912** |
| 6 | 98.737 | 99.650 | 99.287 | 99.075 | 99.700 | 99.525 | 94.588 |
| 7 | 99.025 | 99.675 | 99.213 | 99.013 | 99.575 | 99.513 | 94.662 |
| 8 | 99.050 | 99.150 | 99.100 | 99.150 | **99.750** | 99.588 | 92.463 |
| 9 | **99.138** | 99.675 | **99.362** | 99.188 | 99.438 | 99.388 | 94.237 |
| 10 | 99.037 | **99.800** | 99.138 | 99.100 | 99.388 | 99.537 | 94.263 |

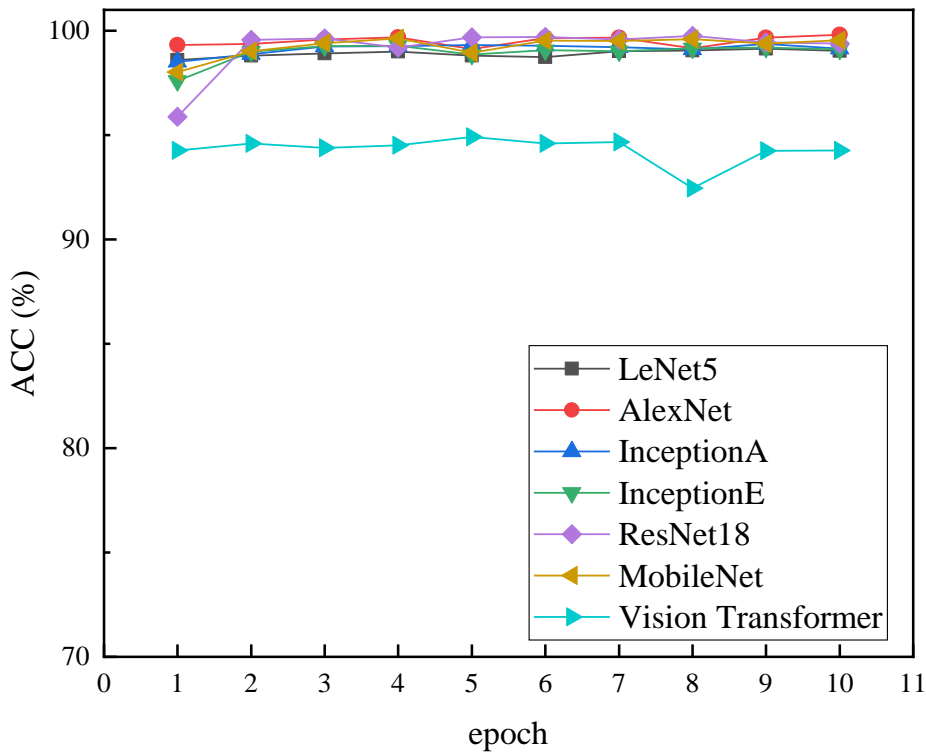

**Fig. 10** Comparison of accuracy of each model before data augmentation

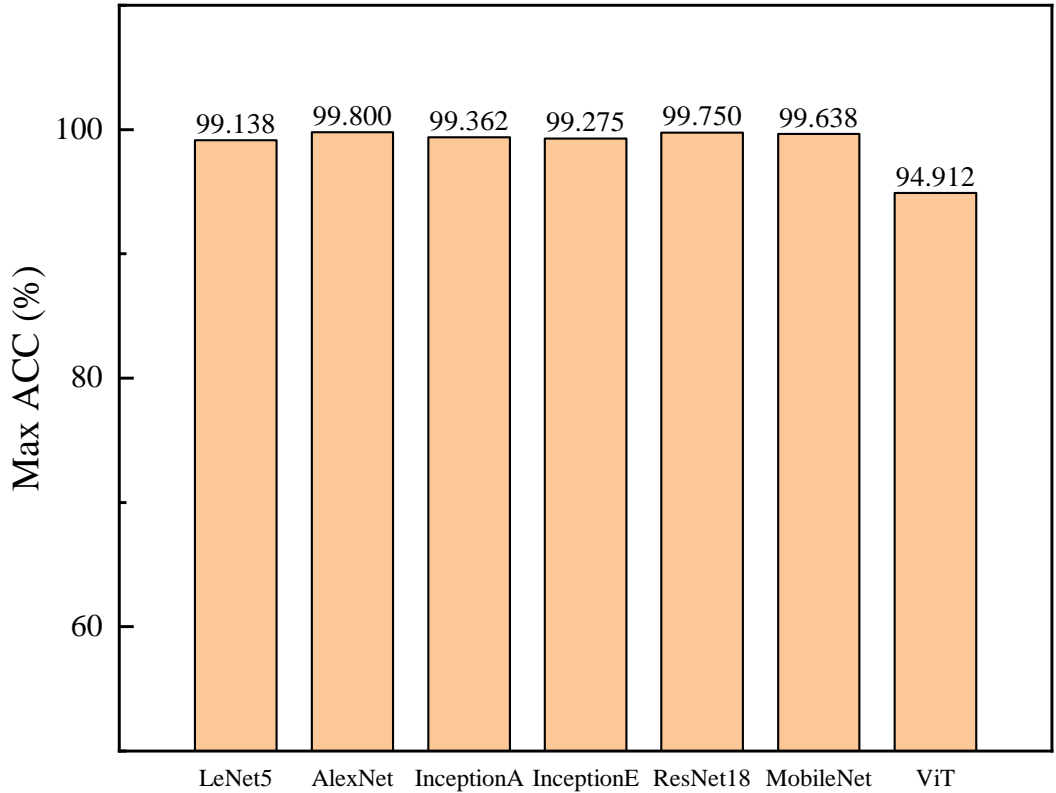

**Fig. 11** Histogram of comparison of the highest accuracy of each model before data augmentation

The quality of the original dataset is very high before the data augmentation. There is no much noise in the dataset. In this case, the accuracy of nearly all models is very high. However, the original dataset is not generic, that is, only very good samples are collected in the dataset. Therefore, after the data augmentation, many worse samples are created and used to train the models.

In this case, the data augmentation operation on the dataset increases the training deviation of the original dataset, resulting in a decrease in accuracy, so as to further analyze the generalization ability and stability of each model. Fig. 13 illustrates the comparison of accuracy rates under 10 epochs before and after data augmentation for the same network model.

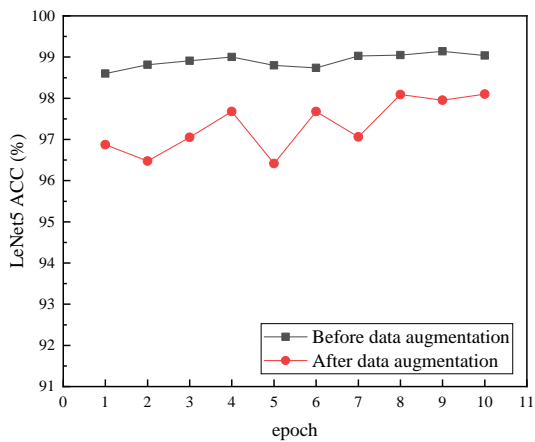

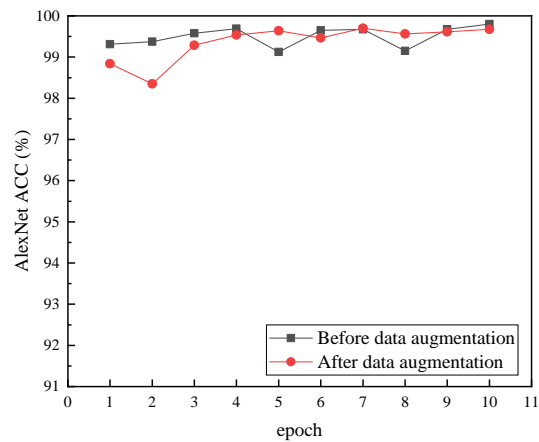

（a）LeNet5　　　　　　　　　　　　　　　（b）AlexNet

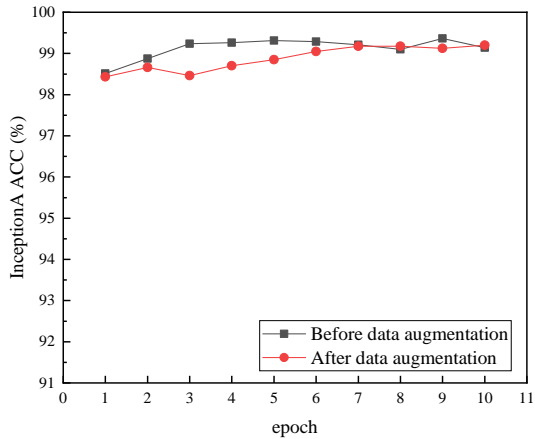
(c) InceptionA

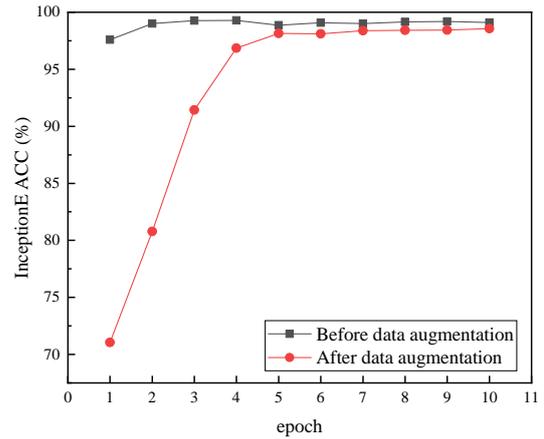
(d) InceptinE

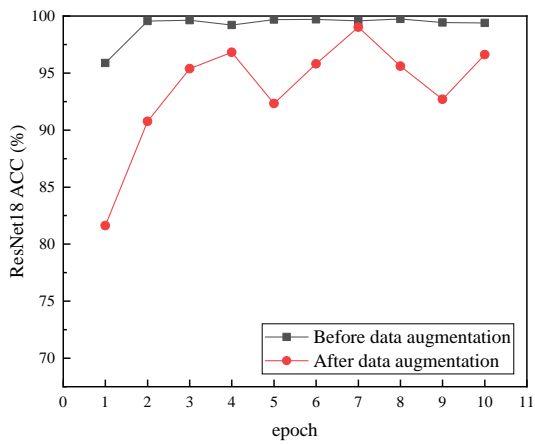
(e) ResNet18

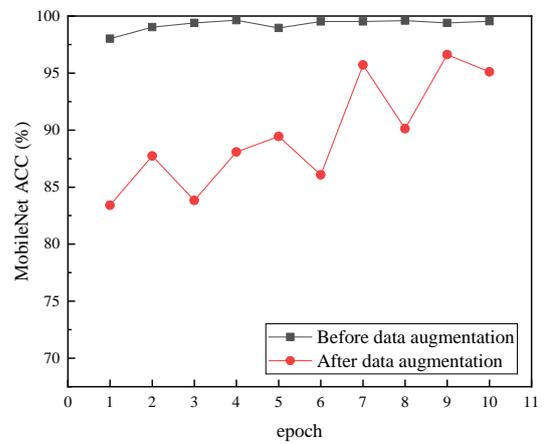
(f) MobileNet

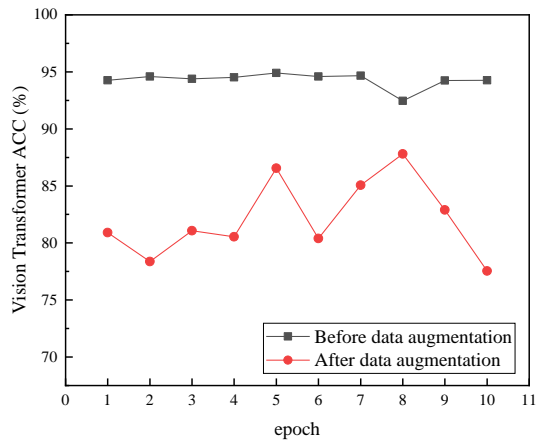
(g) Vision Transformer

**Fig. 12** Comparison of the accuracy of each model before and after data augmentation

From the comparative analysis in Fig. 13, it can be seen that after data augmentation, the accuracy of each model is lower than that before the data augmentation. Among them, the accuracy of AlexNet and InceptionA decreases slightly after data augmentation. Although the accuracy of the ResNet18 fluctuates greatly, the highest accuracy is not much lower than before the data augmentation. It shows that these three types of network models have strong generalization ability. LeNet5, InceptionE, and MobileNet have a large decrease in accuracy after data augmentation, which is lower than other models in generalization ability. The accuracy of Vision Transformer drops the most after data augmentation, indicating that its generalization ability in the field of computer vision is lower than other models. But its accuracy rate can still reach a relatively high level of approximately 87%, indicating that it still has a good development prospect in the field of computer vision. The accuracy comparison of each model under 10 epochs is illustrated in Fig. 14 and Table 4.

Table 4 Accuracy of each model after data augmentation

| epoch | LeNet5 | AlexNet | InceptionA | InceptionE | ResNet18 | MobileNet | VisionTransformer |
|---|---|---|---|---|---|---|---|
| 1 | 96.875 | 98.838 | 93.912 | 71.050 | 86.625 | 83.412 | 80.912 |
| 2 | 96.475 | 98.350 | 98.662 | 80.787 | 90.763 | 87.725 | 78.375 |
| 3 | 97.050 | 99.287 | 98.463 | 91.425 | 95.375 | 83.825 | 81.075 |
| 4 | 97.675 | 99.537 | 98.700 | 96.850 | 96.812 | 88.088 | 80.537 |
| 5 | 96.412 | 99.638 | 98.850 | 98.138 | 92.338 | 89.450 | 86.55 |
| 6 | 97.675 | 99.463 | 99.050 | 98.112 | 95.800 | 86.075 | 80.388 |
| 7 | 97.062 | **99.700** | 99.175 | 98.388 | **99.025** | 95.713 | 85.062 |
| 8 | 98.088 | 99.562 | 99.175 | 98.412 | 95.600 | 90.125 | **87.812** |
| 9 | 97.950 | 99.612 | 99.125 | 98.438 | 92.700 | **96.612** | 82.888 |
| 10 | **98.100** | 99.675 | **99.200** | 98.562 | 96.612 | 95.112 | 77.550 |

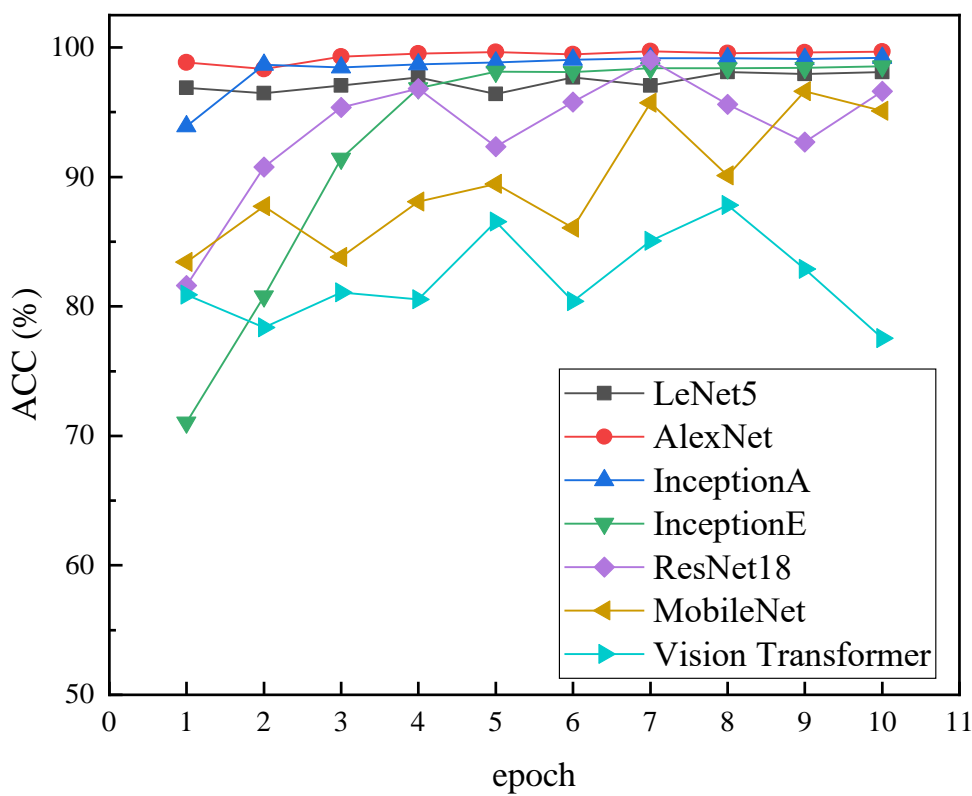

Fig. 13 Comparison of accuracy of each model after data augmentation

From the analysis of Fig. 14 and Table 4, it can be seen that the AlexNet, InceptionA, and ResNet18 still maintain a high level of accuracy after data augmentation, all of which can reach more than 99%. Among them, AlexNet has the highest accuracy, which can reach 99.7 %. After several rounds of training, LeNet and InceptionE can reach an accuracy rate of more than 98%. The accuracy of MobileNet and Vision Transformer is significantly lower than other models. In addition, the accuracy of ResNet18, MobileNet, and Vision Transformer fluctuates greatly, and the model stability is low. The accuracy of InceptionE continues to increase, so it is speculated that it has not reached the highest value of accuracy. After increasing the number of training rounds, the highest accuracy rate of the InceptionE model is 98.800%. Therefore, the highest accuracy of the InceptionE model is 98.800% for comparison with the highest accuracy of other data models. The comparison is illustrated in Fig. 15.

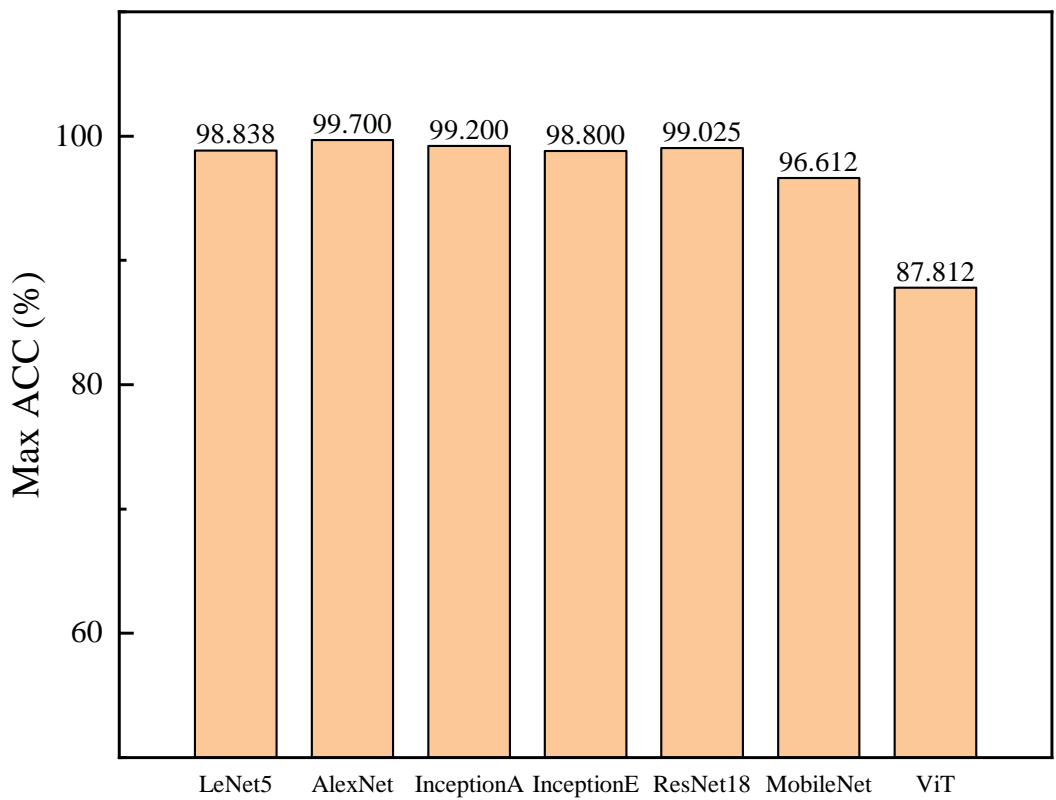

**Fig. 14** Comparison of the highest accuracy of each model after data augmentation

From the analysis of Fig. 15 and Table 4, it can be seen that the accuracy rate of the AlexNet after data augmentation is the highest among the seven models, which is 99.700%. And compared with the accuracy before the data augmentation, the decrease of accuracy is the smallest, which means that when the dataset is relatively small, AlexNet has strong generalization ability and high model stability.

Although the accuracy rates of InceptionA, InceptionE, and LeNet5 have all decreased, and the accuracy rates of LeNet5 and InceptionE have decreased relatively more, their accuracy rates still maintain a high level. In addition, the accuracy of 10 epochs has small fluctuations, the model converges quickly, and a high accuracy rate can be achieved in a short time.

The highest accuracy of ResNet18 is high, reaching more than 99%, but the accuracy of the model fluctuates greatly in 10 epochs. This is due to its Residual Block module, which causes its convergence to slow down, but it avoids the vanishing of the gradient. The accuracy of MobileNet and Transformer has dropped the most, and the accuracy of 10 epochs fluctuates greatly, indicating that the stability of these two models is low, and good training results cannot be achieved for datasets with a large amount of data. But the training time of these two models is relatively short, especially compared with the deep network model such as GoogLeNet, which realizes the lightweight of the model and the improvement of training efficiency.

### 3.3.2 Efficiency

Table 5 Operation timetable for each epoch of each model

| epoch | LeNet5 | AlexNet | InceptionA | InceptionE | ResNet18 | MobileNet | VisionTransformer |
|---|---|---|---|---|---|---|---|
| 1 | 1.34 | 3.40 | 4.76 | 53.39 | 6.62 | 4.31 | 15.02 |
| 2 | 1.33 | 3.32 | 4.30 | 53.24 | 6.73 | 4.25 | 15.02 |
| 3 | 1.33 | 3.32 | 4.30 | 53.24 | 6.70 | 4.21 | 15.05 |
| 4 | 1.34 | 3.31 | 4.28 | 53.30 | 6.69 | 4.37 | 15.00 |
| 5 | 1.33 | 3.33 | 4.29 | 53.25 | 6.70 | 4.51 | 15.00 |
| 6 | 1.33 | 3.33 | 4.26 | 53.25 | 6.69 | 4.54 | 15.01 |
| 7 | 1.33 | 3.35 | 4.31 | 53.26 | 6.71 | 4.54 | 14.99 |
| 8 | 1.34 | 3.33 | 4.29 | 53.29 | 6.78 | 4.54 | 15.00 |
| 9 | 1.32 | 3.36 | 4.25 | 53.26 | 6.69 | 4.53 | 14.98 |
| 10 | 1.32 | 3.32 | 4.31 | 53.27 | 6.72 | 4.52 | 15.00 |

The calculation time of 10 epochs for each model is illustrated in Table 5, and the comparison chart is illustrated in Fig.16. Since the data is enhanced, the training dataset of each model is different in each round of epoch; in this case, the operation time after data augmentation is not comparable. We only compare and analyze the calculation time of each epoch of each model before data augmentation, and make a preliminary judgment on the calculation efficiency of each model.

It can be seen from Fig. 16 that the calculation time of each epoch of each model is not much different, and the calculation time of 10 epochs is evenly distributed. Among those models, the InceptionE has the longest calculation time and the lowest calculation efficiency compared to the other six models; the LeNet5 has the shortest calculation time, which is due to the simplicity of the LeNet5 model and its shallow depth. The Inception module used in the InceptionE model is more complex, which increases the depth and complexity of the model, thus, the calculation time is long, and the calculation efficiency is low. The calculation speed of AlexNet, InceptionA, and MobileNet is similar, approximately 3 ~ 4 minutes. Among them, MobileNet uses a deeply separable convolution module, which greatly improves its calculation speed. Thus, it can maintain a higher calculation efficiency even when the network model is deeper. ResNet18 has a relatively long operation time due to a large number of layers and the deep network model. The operation time of the Vision Transformer is the longest except for InceptionE, indicating that the application of Transformer in the field of computer vision still has a certain optimization space in terms of computational efficiency.

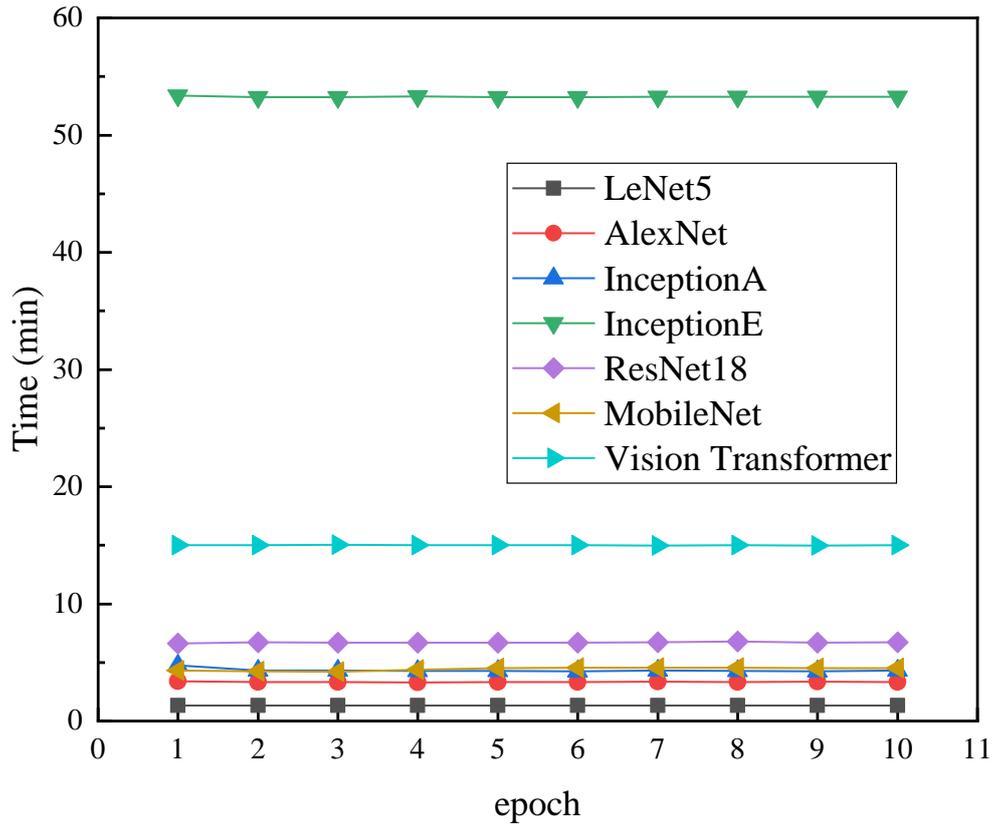

**Fig. 15** Comparison of computing time of each model

## 3.4 Test Results and Analysis of the Refined Deep Learning Models for the Identification of Slope Surface Cracks

### 3.4.1 Comparison of Accuracy Before and After Using Transfer Learning

The small sample dataset of surface cracks in the soil and rock masses is directly used for model training and testing. Taking LeNet5 as an example, the accuracy rate obtained is low, even unable to reach 75%. In this paper, we use the transfer learning idea, combine the training set of the small sample dataset of surface cracks of soil and rock masses into the training set of the large sample dataset of concrete cracks. And we re-use LeNet5 for training; the accuracy rate has been significantly improved, reached 98.094%, an increase of more than 23%. Comparison of accuracy before and after LeNet5 transfer learning is illustrated in Table 6 and Fig. 17.

Table 6 Accuracy comparison table before and after LeNet5 transfer learning

| epoch | 1 | 2 | 3 | 4 | 5 | 6 | 7 | 8 | 9 | 10 |
|---|---|---|---|---|---|---|---|---|---|---|
| Accuracy before transfer learning | 72.360 | 75.000 | 72.500 | 72.500 | 70.000 | 72.500 | 72.500 | 72.500 | 72.500 | 72.500 |
| Accuracy after transfer learning | 96.101 | 97.129 | 97.203 | 97.908 | 97.809 | 97.859 | 98.094 | 98.069 | 97.686 | 97.314 |

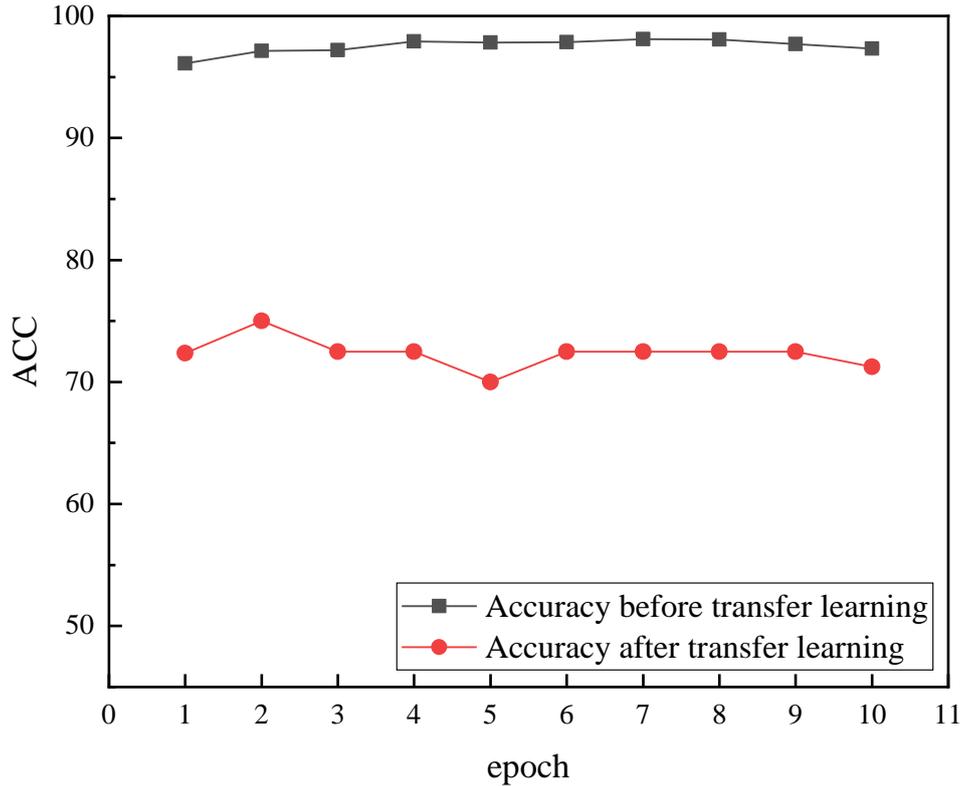

**Fig. 16** Comparison of accuracy before and after LeNet5 transfer learning

### 3.4.2 Comparison of accuracy of each model after transfer learning

The transfer learning method is adopted for the identification of surface cracks of soil and rock masses. Seven network models pre-trained by deep learning: LeNet5, AlexNet, InceptionA, InceptionE, ResNet18, MobileNet, and Vision Transformer are respectively applied to the identification of surface cracks of soil and rock masses. After 10 epochs, the accuracy rate is illustrated in Table 7, and the comparison is illustrated in Figure 18.

Table 7 Comparison of the accuracy of identifying surface cracks of soil and rock masses in each model

| epoch | LeNet5 | AlexNet | InceptionA | InceptionE | ResNet18 | MobileNet | VisionTransformer |
|---|---|---|---|---|---|---|---|
| 1 | 96.101 | 98.899 | 98.886 | 84.196 | 89.719 | 83.824 | 70.520 |
| 2 | 97.129 | 98.837 | 98.824 | 85.619 | 94.356 | 82.599 | 79.498 |
| 3 | 97.203 | 99.468 | 98.280 | 96.832 | 95.099 | 86.572 | 84.567 |
| 4 | 97.908 | 99.319 | 98.812 | 97.834 | 83.317 | 91.064 | 82.587 |
| 5 | 97.809 | 99.233 | 98.725 | 98.069 | 95.532 | 91.708 | 83.688 |
| 6 | 97.859 | 99.433 | 98.651 | 98.082 | 90.149 | 87.698 | 81.002 |
| 7 | **98.094** | 99.381 | 98.960 | 98.403 | 94.406 | 90.916 | 76.535 |
| 8 | 98.069 | **99.666** | **98.973** | 98.391 | **96.609** | **97.079** | **89.579** |
| 9 | 97.686 | 99.418 | 98.948 | **98.490** | 91.955 | 96.782 | 85.594 |
| 10 | 97.314 | 99.604 | 98.861 | 98.428 | 86.399 | 93.181 | 77.822 |

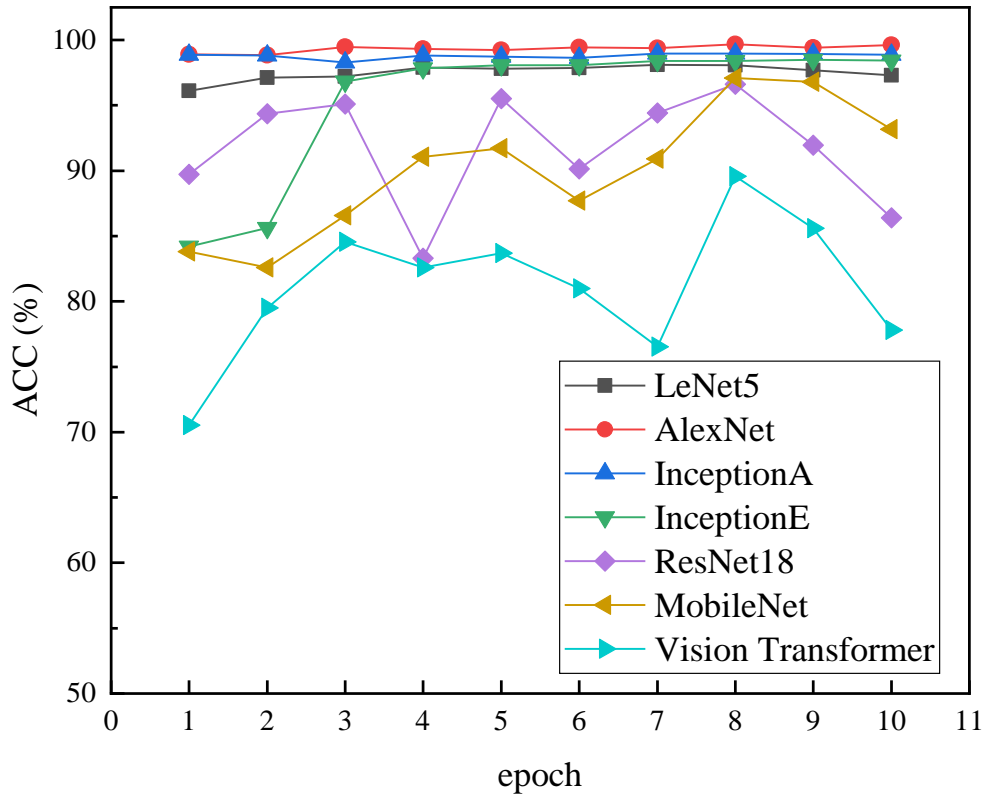

Fig. 17 Comparison of the accuracy rate of identifying the surface cracks of soil and rock masses of each model

From the analysis of Fig. 18 and Table 7, it can be seen that the identification of surface cracks of soil and rock masses can reach a higher accuracy after transfer learning. Except for Vision Transformer, the accuracy of the other six models can reach more than 95%, and the highest accuracy of the Vision Transformer can also reach 89.597%, which is close to 90%. Therefore, the application of transfer learning has a better effect on the identification of surface cracks of soil and rock masses.

In order to make the training test results obtained after inserting the soil and rock crack data training set into the original dataset more real, the training set is processed with data augmentation. The accuracy curve for the identification of surface cracks of soil and rock masses after transfer learning is basically the same as the original dataset curve shape that is enhanced before transfer learning. LeNet5, AlexNet, InceptionA, and InceptionE are all accurate and stable, while ResNet18, MobileNet, and Vision Transformer have relatively high volatility. It can be seen that each model maintains its original characteristics after transfer learning and does not cause major changes due to transfer learning.

### 3.4.3 Comparative Analysis of the Accuracy of the Refined Models and the Pre-trained Models

Seven pre-trained deep learning network models: LeNet5, AlexNet, InceptionA, InceptionE, ResNet18, MobileNet, and Vision Transformer are employed for the identification of surface cracks of soil and rock masses. Compared with the identification of concrete cracks conducted by the original dataset training, the accuracy comparison analysis chart under 10 epochs is illustrated in Fig. 19.

Take the highest accuracy rate of each model when using the original dataset and compare the highest accuracy rate of each model after using the surface cracks of soil and rock masses dataset for transfer learning, and a comparison is illustrated in Fig. 20.

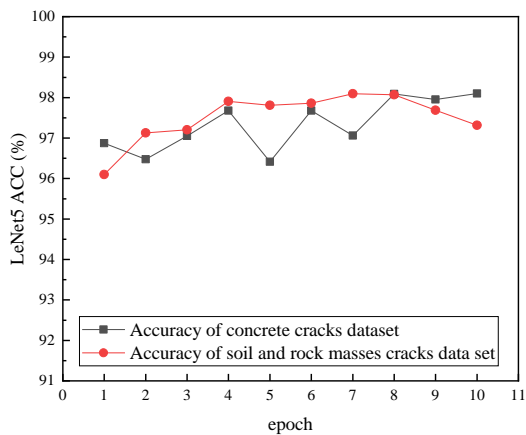

（a）LeNet5

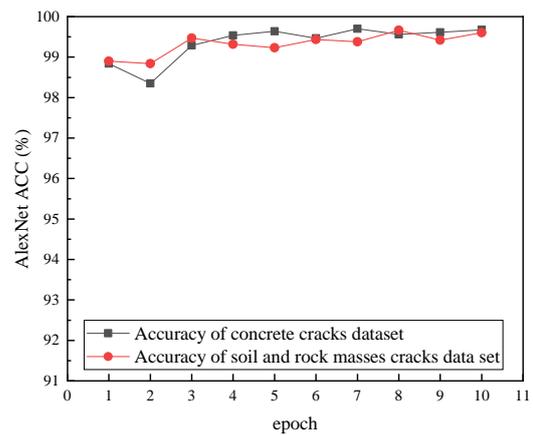

（b）AlexNet

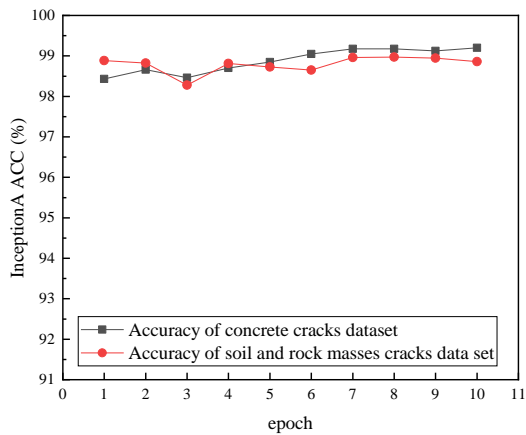

（c）InceptionA

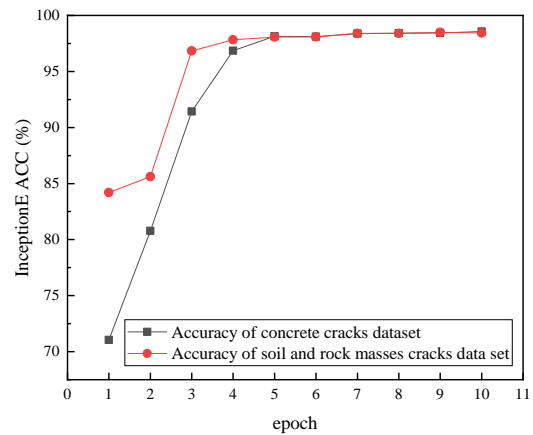

（d）InceptionE

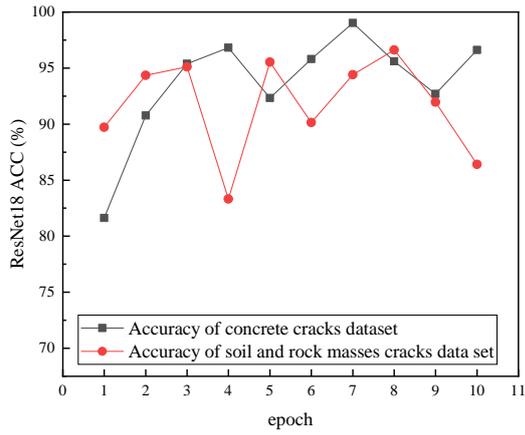

（f）ResNet18

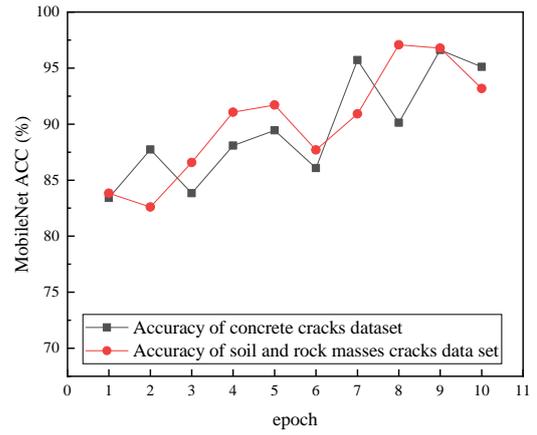

（f）MobileNet

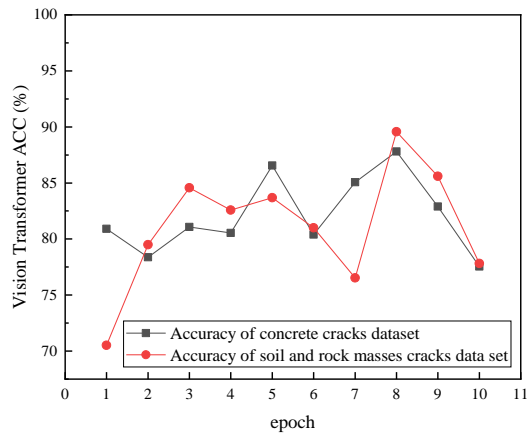

（g）Vision Transformer

**Fig. 18** Comparison of the accuracy of the identification of surface cracks of soil and rock masses and concrete cracks of each model

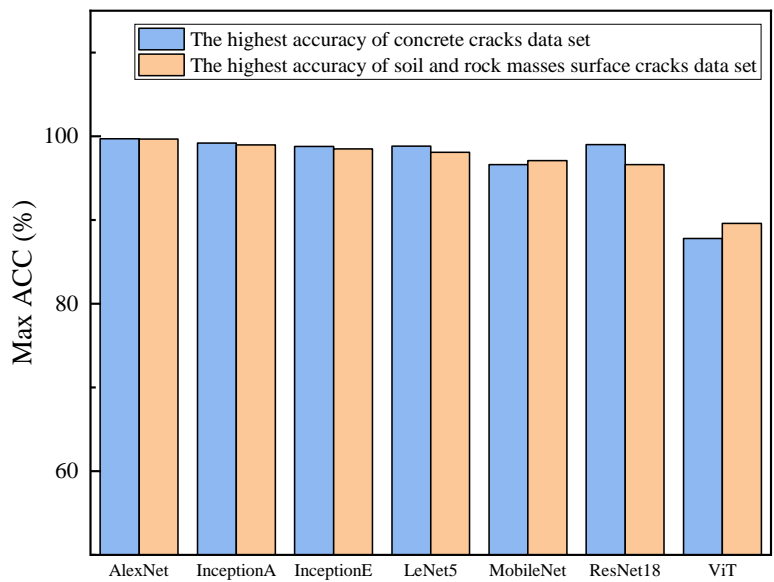

**Fig. 19** Histogram of the highest accuracy comparison between the identification of surface cracks of soil and rock masses and concrete cracks of each model

From the analysis in Fig. 20, it can be seen that when the identification of slope cracks deep learning refined models are used to identify surface cracks of soil and rock masses, the accuracy rate basically maintains the accuracy level of the original pre-trained models. The accuracy of MobileNet and Vision Transformer has even increased, which proves that the generalization ability of the deep learning refined models has been improved. This shows that it is feasible to employ transfer learning between the slope surface cracks of soil and rock masses and concrete cracks, and the training effect is better. Among the seven models, the AlexNet has the highest accuracy in both the original dataset and the soil and rock crack dataset, further indicating that the AlexNet is suitable for the identification of slope surface cracks. The accuracy of MobileNet and Vision Transformer did not decrease but increased after using transfer learning which indicates the strong adaptability of MobileNet and Vision Transformer to the identification of soil and rock masses slope surface cracks. And the MobileNet and Vision Transformer have good development potential in the application for the identification of soil and rock masses slope surface cracks in the future.

# 4 Discussion

In this paper, we present a deep transfer learning framework for identifying the slope surface cracks. The following section will discuss the advantages and some shortcomings of the presented framework, as well as prospects for future work.

## 4.1 Advantages and Applicability

By exploiting the idea of transfer learning, the large sample dataset of concrete cracks (a total of 40,000 images) is combined with the small sample surface cracks of soil and rock masses dataset (a total of 400 images) which is collected, sorted, and annotated by ourselves. The pre-trained deep learning models are employed for the identification of surface cracks of soil and rock masses. A high accuracy rate has been achieved. And the limitation of less dataset of slope surface cracks of soil and rock masses is overcome. The massive amount of data required for deep learning model training is also reduced greatly. Therefore, it is hoped to build a refined model of deep learning that can be deployed on UAVs for UAV aerial surveys to identify slope surface cracks and be applied to geological disaster monitoring and early warning.

Taking LeNet5 as an example. Before transfer learning, when only the small sample dataset is used, the highest accuracy rate can only reach 75%. After transfer learning, the large sample dataset and the small sample dataset are combined, the highest accuracy rate of the identification of slope cracks has been significantly improved, reached 98.094%, an increase of more than 23%. The massive data required for the identification of slope surface cracks is significantly reduced.

The flexibility of the model is related to the depth of the model. The more layers of the model, the deeper and the stronger the flexibility of the model. However, the more layers of the model, the more potential features of the dataset are extracted. When faced with a small-scale dataset, models with many layers are prone to overfitting. Deep learning models with few layers and shallow depth can achieve better learning effects. In this paper, the dataset is small. The combination of large sample dataset and small sample dataset only has 40,400 images. Therefore, the AlexNet model and LeNet model with a small number of layers have a higher accuracy rate, and the accuracy of ResNet18 and Vision Transformer models with more layers is lower.

## 4.2 Shortcomings

The refined models in this paper can only identify slope surface cracks, but the location, width, length, and depth of the slope surface cracks are still uncertain, and further optimization is needed. Moreover, In the construction of the dataset for the identification of slope cracks in this paper, the dataset is still small. For some models that are suitable for large sample datasets, good training and test results cannot be obtained, such as Vision Transformer.

## 4.3 Future Work

(1) Due to the lack of image data of surface cracks of soil and rock masses, there is little research work focusing on the deep learning of surface cracks of soil and rock masses. Therefore, it is very important to collect the image data of surface cracks of soil and rock masses and establish the dataset of surface cracks of soil and rock masses. In the future, UAV aerial surveys can be used to collect crack images in the study area, and image cropping can be carried out through relevant codes, so as to establish a dataset of surface cracks of soil and rock masses in the study area. In addition, surface cracks of soil and rock masses dataset sharing network platform can also be established, and users can upload crack pictures in real-time to share with other researchers, thereby establishing a huge soil and rock crack database, which is convenient for researchers to study and learn about surface cracks of soil and rock masses.

(2) With the development of deep learning, there will be more deep neural networks with high accuracy and excellent performance in the future. This paper only conducts related research on the identification of cracks. In the future, more detailed research will be conducted on the location, length, width, and depth of the cracks. A deep learning model of slope crack detection suitable for UAV aerial surveys will be established, so as to realize the practical application of deep learning in the field for the identification of slope cracks.

(3) Currently, there are few deep learning models designed specifically for the identification of slope surface cracks. Therefore, future research work will focus on the design of related deep learning models for the identification of slope cracks. In addition, the deep learning model is deployed on the UAV to monitor the high and steep slopes and other locations that are prone to geohazards and difficult to detect on-site cracks. Real-time monitoring of the surface cracks of soil and rock masses can be carried out to timely monitor and early warning of geological disasters on high and steep slopes to ensure the safety of people's lives and property.

# 5 Conclusion

In this paper, we propose a deep transfer learning framework to effectively and efficiently identify slope surface cracks for the sake of early monitoring and warning of geohazards such as landslides. The essential idea is to employ transfer learning by training (a) the large sample dataset of concrete cracks and (b) the small sample dataset of soil and rock masses cracks. In the proposed framework, (1) pre-trained cracks identification models are constructed based on the large sample dataset of concrete cracks; (2) refined cracks identification models are further constructed based on the small sample dataset of soil and rock masses cracks. To evaluate the effectiveness of the proposed framework, the accuracy of the deep learning models is tested. The results show that: (1) AlexNet has the highest accuracy and computational efficiency, proves that the classic AlexNet model is effective in identifying slope cracks. (2) Each model has achieved much higher accuracy by employing the idea of transfer learning. The accuracy of LeNet is approximately 23% higher than using the small sample dataset. The massive data required for the identification of slope surface cracks is significantly reduced.

Future work is planned to achieve better results for a sufficient-sized dataset. The deep transfer learning framework can be deployed on UAV aerial surveys, serving the detection of surface cracks on high and steep slopes. And the deep transfer learning framework has important engineering significance to realize the timely monitoring, so as to carry out the prevention of geological disasters.